\documentclass[sigconf]{acmart}

\usepackage{siunitx}
\usepackage{enumitem}
\usepackage{indentfirst}
\usepackage{xurl}

\usepackage[acronym,shortcuts]{glossaries}
\newacronym{ARMD}{ARMD}{age-related macular degeneration}
\newacronym{FDR}{FDR}{false-discovery rate}
\newacronym{NFB}{NFB}{nerve fiber bundle}
\newacronym{RGC}{RGC}{retinal ganglion cell}
\newacronym{RP}{RP}{retinitis pigmentosa}
\newacronym{SPV}{SPV}{simulated prosthetic vision}
\newacronym{UCSB}{UCSB}{University of California, Santa Barbara}
\newacronym{VPU}{VPU}{video processing unit}
\newacronym{VR}{VR}{virtual reality}

\AtBeginDocument{%
  \providecommand\BibTeX{{%
    \normalfont B\kern-0.5em{\scshape i\kern-0.25em b}\kern-0.8em\TeX}}}

\setcopyright{acmcopyright}
\copyrightyear{2021}
\acmYear{2021}
\acmDOI{10.1145/1122445.1122456}

\acmConference[AHs '21]{Augmented Humans '21}{February 22--24, 2021}{Online}
\acmBooktitle{Augmented Humans '21}
\acmPrice{15.00}
\acmISBN{978-1-4503-XXXX-X/18/06}

\begin{document}

\title{Deep Learning--Based Scene Simplification for Bionic Vision}



\author{Nicole Han}
\affiliation{
    \institution{University of California}
    \city{Santa Barbara}
    \state{CA}
    \country{USA}
}
\email{xhan01@ucsb.edu}

\author{Sudhanshu Srivastava}
\authornote{Both authors contributed equally to this research.}
\affiliation{
    \institution{University of California}
    \city{Santa Barbara}
    \state{CA}
    \country{USA}
}
\email{sudhanshu@ucsb.edu}

\author{Aiwen Xu}
\authornotemark[1]
\affiliation{
    \institution{University of California}
    \city{Santa Barbara}
    \state{CA}
    \country{USA}
}
\email{aiwenxu@ucsb.edu}

\author{Devi Klein}
\affiliation{
    \institution{University of California}
    \city{Santa Barbara}
    \state{CA}
    \country{USA}
}
\email{dklein@ucsb.edu}

\author{Michael Beyeler}
\orcid{0000-0001-5233-844X}
\affiliation{
    \institution{University of California,}
    \city{Santa Barbara}
    \state{CA}
    \country{USA}
}
\email{mbeyeler@ucsb.edu}

\renewcommand{\shortauthors}{Han et al.}

\begin{abstract}
\noindent Retinal degenerative diseases cause profound visual impairment in more than 10 million people worldwide, and retinal prostheses are being developed to restore vision to these individuals.
Analogous to cochlear implants, these devices electrically stimulate surviving retinal cells to evoke visual percepts (phosphenes).
However, the quality of current prosthetic vision is still rudimentary.
Rather than aiming to restore ``natural'' vision, there is potential merit in borrowing state-of-the-art computer vision algorithms as image processing techniques to maximize the usefulness of prosthetic vision.
Here we combine deep learning--based scene simplification strategies with a psychophysically validated computational model of the retina to generate realistic predictions of simulated prosthetic vision, and measure their ability to support scene understanding of sighted subjects (virtual patients) in a variety of outdoor scenarios.
We show that object segmentation may better support scene understanding than models based on visual saliency and monocular depth estimation.
In addition, we highlight the importance of basing theoretical predictions on biologically realistic models of phosphene shape.
Overall, this work has the potential to drastically improve the utility of prosthetic vision for people blinded from retinal degenerative diseases.

\end{abstract}



\begin{teaserfigure}
  \centering
  \includegraphics[width=0.92\textwidth]{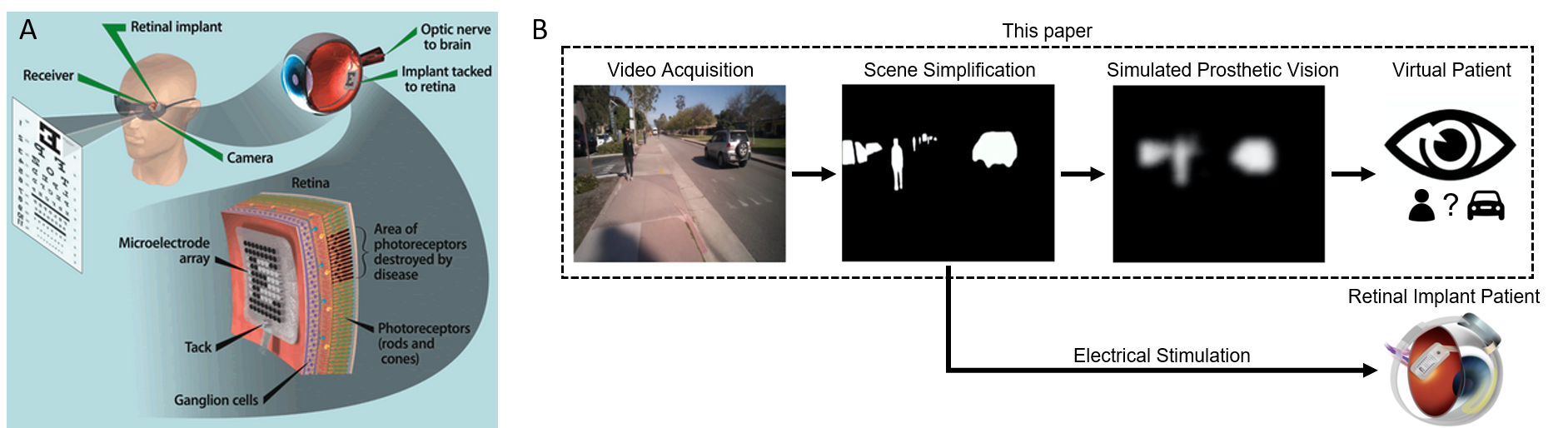}
  \caption{Retinal implant (`bionic eye') for restoring vision to people with visual impairment.
  \emph{A)} Light captured by a camera is transformed into electrical pulses delivered through a microelectrode array to stimulate the retina (adapted with permission from \cite{weiland_biomimetic_2005}).
  \emph{B)} To create meaningful artificial vision, we explored deep learning--based scene simplification as a preprocessing strategy for retinal implants (reproduced from \href{https://doi.org/10.6084/m9.figshare.13652927}{doi:10.6084/m9.figshare.13652927} under CC-BY 4.0).
  As a proof of concept, we used a neurobiologically inspired computational model to generate realistic predictions of \acf{SPV}, and asked sighted subjects (i.e., virtual patients) to identify people and cars in a novel \acs{SPV} dataset of natural outdoor scenes.
  In the future, this setup may be used as input to a real retinal implant.}
  \Description{Deep learning--based scene simplification for bionic vision.}
  \label{fig:teaser}
\end{teaserfigure}

\maketitle

\pagebreak
\section{Introduction}

Retinal degenerative diseases such as \ac{RP} and \ac{ARMD} lead to a gradual loss of photoreceptors in the eye that may cause profound visual impairment in more than 10 million people worldwide.
Analogous to cochlear implants, retinal neuroprostheses (also known as the \emph{bionic eye}, Fig.~\ref{fig:teaser}A) aim to restore vision to these individuals by electrically stimulating surviving retinal cells to evoke neuronal responses that are interpreted by the brain as visual percepts (\emph{phosphenes}).
Existing devices generally provide an improved ability to localize high-contrast objects, navigate, and perform basic orientation tasks \cite{ayton_update_2020}.
Future neural interfaces will likely enable applications such as controlling complex robotic devices, extending memory, or augmenting natural senses with artificial inputs \cite{fernandez_development_2018}.

However, despite recent progress in the field, there are still several limitations affecting the possibility to provide useful vision in daily life \cite{beyeler_learning_2017}.
Interactions between the device electronics and the underlying neurophysiology of the retina have been shown to lead to distortions that can severely limit the quality of the generated visual experience \cite{fine_pulse_2015,beyeler_model_2019}.
Other challenges include how to improve visual acuity, enlarge the field-of-view, and reduce a complex visual scene to its most salient components through image processing.

Rather than aiming to restore ``natural'' vision, there is potential merit in borrowing computer vision algorithms as preprocessing techniques to maximize the usefulness of bionic vision.
Whereas edge enhancement and contrast maximization are already routinely employed by current devices, relatively little work has explored the extraction of high-level scene information.

To address these challenges, we make three contributions:
\begin{enumerate}[topsep=0pt,itemsep=-1ex,partopsep=0pt,parsep=1ex,leftmargin=14pt,label=\roman*.]
    \item We adopt state-of-the-art computer vision algorithms to explore deep learning--based scene simplification as a preprocessing strategy for bionic vision.
    \item Importantly, we use an established and psychophysically validated computational model of bionic vision to generate realistic predictions of \acf{SPV}.
    \item We systematically evaluate the ability of these algorithms to support scene understanding with a user study focused on a novel dataset of natural outdoor scenes.
\end{enumerate}




\section{Background}
\label{sec:background}

Retinal implants are currently the only FDA-approved technology to treat blinding degenerative diseases such as \ac{RP} and \ac{ARMD}.
Most current devices acquire visual input via an external camera and perform edge extraction or contrast enhancement via an external \ac{VPU}, before sending the signal through wireless coils to a microstimulator implanted in the eye or the brain (see Fig.~\ref{fig:teaser}A).
This device receives the information, decodes it and stimulates the visual system with electrical current, ideally resulting in artificial vision.
Two devices are already approved for commercial use: Argus II (60 electrodes, Second Sight Medical Products, Inc., \cite{luo_argusr_2016}) and Alpha-IMS (1500 electrodes, Retina Implant AG, \cite{stingl_artificial_2013}). In addition, PRIMA (378 electrodes, Pixium Vision, \cite{lorach_photovoltaic_2015}) has started clinical trials, with others to follow shortly \cite{ayton_first--human_2014,ferlauto_design_2018}.

However, a major outstanding challenge in the use of these devices is translating electrode stimulation into a code that the brain can understand.
A common misconception is that each electrode in the grid can be thought of as a `pixel' in an image \cite{dagnelie_real_2007,chen_simulating_2009,lui_transformative_2011,perez-yus_depth_2017,sanchez-garcia_indoor_2019}, and most retinal implants linearly translate the grayscale value of a pixel in each video frame to a current amplitude of the corresponding electrode in the array \cite{luo_argusr_2016}.
To generate a complex visual experience, the assumption then is that one simply needs to turn on the right combination of pixels.

In contrast, a growing body of evidence suggests that individual electrodes do not lead to the perception of isolated, focal spots of light \cite{fine_pulse_2015,beyeler_model_2019,erickson-davis_what_2020}.
Although consistent over time, phosphenes vary drastically across subjects and electrodes \cite{luo_long-term_2016,beyeler_model_2019} and often fail to assemble into more complex percepts \cite{rizzo_perceptual_2003,wilke_electric_2011}.
Consequently, retinal implant users do not see a perceptually intelligible world \cite{erickson-davis_what_2020}.

\begin{figure}[b!]
\centering
\includegraphics[width=\columnwidth]{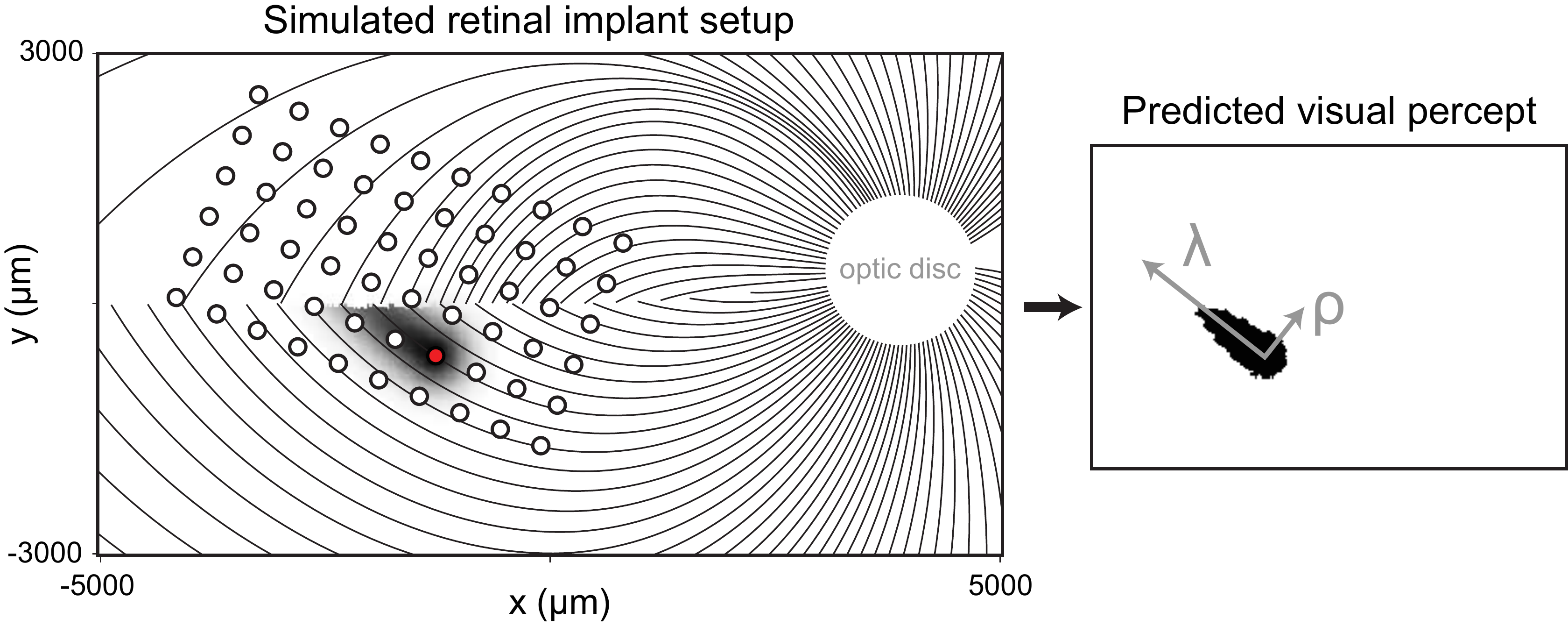}
\caption{A simulated map of retinal \acsp{NFB} (\emph{left})
    can account for visual percepts (\emph{right}) 
    elicited by retinal implants (reprinted with permission from \cite{beyeler_model-based_2019}).
    \emph{Left}: Electrical stimulation (red circle) of a \acs{NFB} (black lines) could activate retinal ganglion cell bodies peripheral to the point of stimulation,
    leading to tissue activation (black shaded region) elongated
    along the \acs{NFB} trajectory away from the optic disc (white circle).
    \emph{Right}: The resulting visual percept appears elongated as well; its shape can be described by two parameters, $\lambda$ (spatial extent along the \acs{NFB} trajectory) and $\rho$ (spatial extent perpendicular to the \acs{NFB}).
    See Ref.~\cite{beyeler_model-based_2019} for more information.
}
\label{fig:axonmap}
\end{figure}

A recent study demonstrated that the shape of a phosphene generated by a retinal implant depends on the retinal location of the stimulating electrode \cite{beyeler_model_2019}.
Because \acp{RGC} send their axons on highly stereotyped pathways to the optic nerve, an electrode that stimulates nearby axonal fibers would be expected to antidromically activate \ac{RGC} bodies located peripheral to the point of stimulation, leading to percepts that appear elongated in the direction of the underlying \ac{NFB} trajectory (Fig.~\ref{fig:axonmap}, \emph{right}).
Ref.~\cite{beyeler_model_2019} used a simulated map of \acp{NFB} in each patient's retina to accurately predict phosphene shape, by assuming that an axon's sensitivity to electrical stimulation:
\begin{enumerate}[topsep=0pt,itemsep=-1ex,partopsep=0pt,parsep=1ex,leftmargin=14pt,label=\roman*.]
    \item decays exponentially with decay constant $\rho$ as a function of distance from the stimulation site,
    \item decays exponentially with decay constant $\lambda$ as a function of distance from the cell body, measured as axon path length.
\end{enumerate}

As can be seen in Fig.~\ref{fig:axonmap} (\emph{left}), electrodes near the horizontal meridian are predicted to elicit circular percepts, while other electrodes are predicted to produce elongated percepts that will differ in angle based on whether they fall above or below the horizontal meridian.
In addition, the values of $\rho$ and $\lambda$ dictate the size and elongation of elicited phosphenes, respectively, which may drastically affect visual outcomes.
Understanding the qualitative experience associated with retinal implants and finding ways to improve is therefore indispensable to the development of visual neuroprostheses and related vision augmentation technologies.

\section{Related Work}
\label{sec:related_work}

Most retinal implants are equipped with an external \ac{VPU} that is capable of applying simple image processing techniques to the video feed in real time.
In the near future, these techniques may include deep learning--based algorithms aimed at improving a patient's scene understanding.

Based on this premise, researchers have developed various image optimization strategies, and assessed their performance by having sighted observers (i.e., \emph{virtual patients}) conduct daily visual tasks under \ac{SPV} \cite{boyle_region--interest_2008,dagnelie_real_2007,al-atabany_improved_2010,li_image_2018,mccarthy_mobility_2014,vergnieux_simplification_2017}.
Simulation allows a wide range of computer vision systems to be developed and tested without requiring implanted devices.

\Ac{SPV} studies suggest that one benefit of image processing may be to provide an importance mapping that can aid scene understanding; that is, to enhance certain image features or regions of interest, at the expense of discarding less important or distracting information \cite{boyle_region--interest_2008,al-atabany_improved_2010,horne_semantic_2016,sanchez-garcia_indoor_2019}.
This limited compensation may be significant to retinal prosthesis patients carrying out visual tasks in daily life.

One of the most commonly explored strategies is to highlight visually salient information in the scene.
In biologically-inspired models, visual saliency is often defined as a \emph{bottom-up} process that highlights regions whose low-level visual attributes (e.g., color, contrast, motion) may differ from their immediate surroundings.
Early work used this approach to build a visual saliency map whose salient regions coincided with the regions gazed at by human subjects when looking at images \cite{parikh_saliency-based_2010}.
More recent research showed that saliency was able to improve eye-hand coordination \cite{li_real-time_2017}, obstacle avoidance \cite{stacey_salient_2011}, object detection \cite{weiland_smart_2012}, and object recognition \cite{li_image_2018,wang_image_2016}.
However, saliency prediction improved markedly with the advent of deep learning models, which are commonly trained on human eye movement data to predict an observer's fixation locations while freely-viewing a set of images.
The current state-of-the-art in saliency prediction is DeepGaze II \cite{kummerer_deepgaze_2016}, a probabilistic model that uses transfer learning from VGG-19 pre-trained on the SALICON dataset. 
DeepGaze has yet to be applied to the field of bionic vision.

Current retinal prostheses are implanted in only one eye, and thus are unable to convey binocular depth cues.
Previous work has therefore explored the possibility of obtaining depth information through additional peripherals, such as an RGB-D sensor, and studied behavioral performance of virtual patients typically navigating an obstacle course under \ac{SPV}.
For example, Ref.~\cite{perez-yus_depth_2017} used depth cues to generate a simplified representation of the ground to indicate the free space within which virtual patients could safely walk around.
Depth cues were also shown to help avoid nearby obstacles that are notoriously hard to detect with other computer vision algorithms, such as branches hanging from a tree \cite{lieby_substituting_2011}.
Ref.~\cite{mccarthy_mobility_2014} used depth to increase the contrast of object boundaries and showed that this method reduced the number of collisions with ground obstacles.
In addition, retinal prosthesis patients were shown to benefit from distance information provided by a thermal sensor when trying to avoid nearby obstacles and people \cite{sadeghi_thermal_2019}.
However, recent advances in deep learning enable the estimation of relative depth from single monocular images, thereby eliminating the need of external depth sensors and peripherals. One of the most promising deep neural networks is monodepth2 \cite{godard_digging_2019}, which uses a self-supervised method to estimate per-pixel monocular depth.
Deep learning--based depth estimation has yet to be applied to the field of bionic vision.


Finally, recent advances in semantic segmentation have found application in bionic vision to simplify the representation of both outdoor scenes \cite{horne_semantic_2016} and indoor scenes \cite{sanchez-garcia_indoor_2019}.
The latter study combined semantic and structural image segmentation to build a schematic representation of indoor environments, which was then shown to improve object and room identification in a \ac{SPV} task \cite{sanchez-garcia_indoor_2019}. 

However, a common limitation of all the above studies is that their prosthetic vision simulation assumed that phosphenes are small, isolated, independent light sources.
It is therefore unclear how their findings would translate to real retinal prosthesis patients, whose phosphenes are large, elongated, and often fail to assemble into more complex percepts \cite{rizzo_perceptual_2003,wilke_electric_2011,beyeler_model_2019,erickson-davis_what_2020}.
In addition, since the above algorithms were developed in isolation and tested on different behavioral tasks, a side-by-side comparison of their ability to aid scene understanding is still lacking.

To address these challenges, we used a neurobiologically inspired computational model of bionic vision to generate realistic predictions of \ac{SPV}, and applied it to several state-of-the-art computer vision algorithms that might be used to aid scene understanding.
To allow for a fair comparison between algorithms, we asked virtual patients to make perceptual judgments about natural outdoor scenes, and assessed their performance using objective metrics as we systematically varied a number of model parameters.

\section{Methods}

\begin{figure*}
    \centering
    \includegraphics[width=\textwidth]{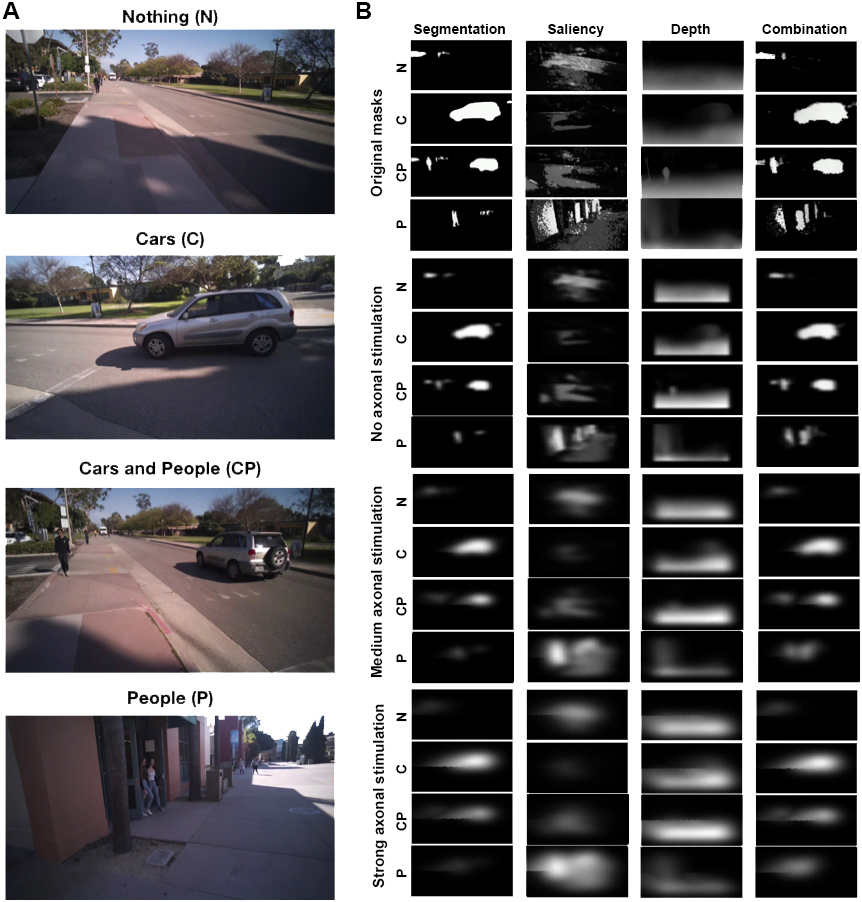}
    \caption{Example frames of the \acf{SPV} dataset.
    \emph{A)} Example frames containing: neither people nor cars (N), cars only (C), both cars and people (CP), and people only (P).
    \emph{B)} Same example frames after being processed with different scene simplification strategies (columns) and \acs{SPV} of a $32 \times 32$ electrode array with different phosphene sizes and elongations (rows).
    Simulations are shown for the original masks (no \acs{SPV}), small phosphenes with no axonal stimulation ($\rho=\SI{100}{\micro\meter}, \lambda=\SI{0}{\micro\meter}$), medium-sized phosphenes with intermediate axonal stimulation ($\rho=\SI{300}{\micro\meter}, \lambda=\SI{1000}{\micro\meter}$), and large phosphenes with strong axonal stimulation ($\rho=\SI{500}{\micro\meter}, \lambda=\SI{5000}{\micro\meter}$).
    Phosphene size and elongation drastically affect \acs{SPV} quality, but previous work often ignored these parameters in their predictions.}
    \label{fig:stimuli}
\end{figure*}

Following the workflow outlined in Fig.~\ref{fig:teaser}B, we created \ac{SPV} videos of various outdoor scenes captured by a head-mounted camera (Section~\ref{sec:stimuli}).
We first processed the raw videos with one of four scene simplification strategies based on state-of-the-art computer vision algorithms (Section~\ref{sec:strategies}).
We then fed the preprocessed videos into a prosthetic vision simulator to simulate the artificial vision likely to be experienced by different retinal prosthesis patients (Section~\ref{sec:SPV}).
Example frames of the resulting \ac{SPV} videos can be seen in Fig.~\ref{fig:stimuli}.
Finally, we conducted a user study to evaluate how well the resulting \ac{SPV} videos could support scene understanding in a variety of outdoor scenarios (Section~\ref{sec:virtual_patients}).


\subsection{Visual Stimuli}
\label{sec:stimuli}

Stimuli consisted of $16$ first-person videos (each \SI{5}{\second} long) recorded on the \ac{UCSB} campus using head-mounted Tobii Pro Glasses 2.
All videos were recorded outdoors in broad daylight, and were aimed at capturing scenarios that are relevant for orientation and mobility of a retinal prosthesis patient (e.g., walking on a sidewalk, crossing a street, strolling through a park).
The final dataset was carefully assembled so as to uniformly cover a variety of conditions. That is, four videos did not include any people or cars; four videos had one or more person present; four videos had one or more cars present; and four videos had both people and cars present.

The raw dataset is publicly available as part of the Supplementary Material (see Section~\ref{sec:data_availability}).

\newpage
\vfill

\vfill
\clearpage

\begin{figure*}[!t]
    \centering
    \includegraphics[width=\textwidth]{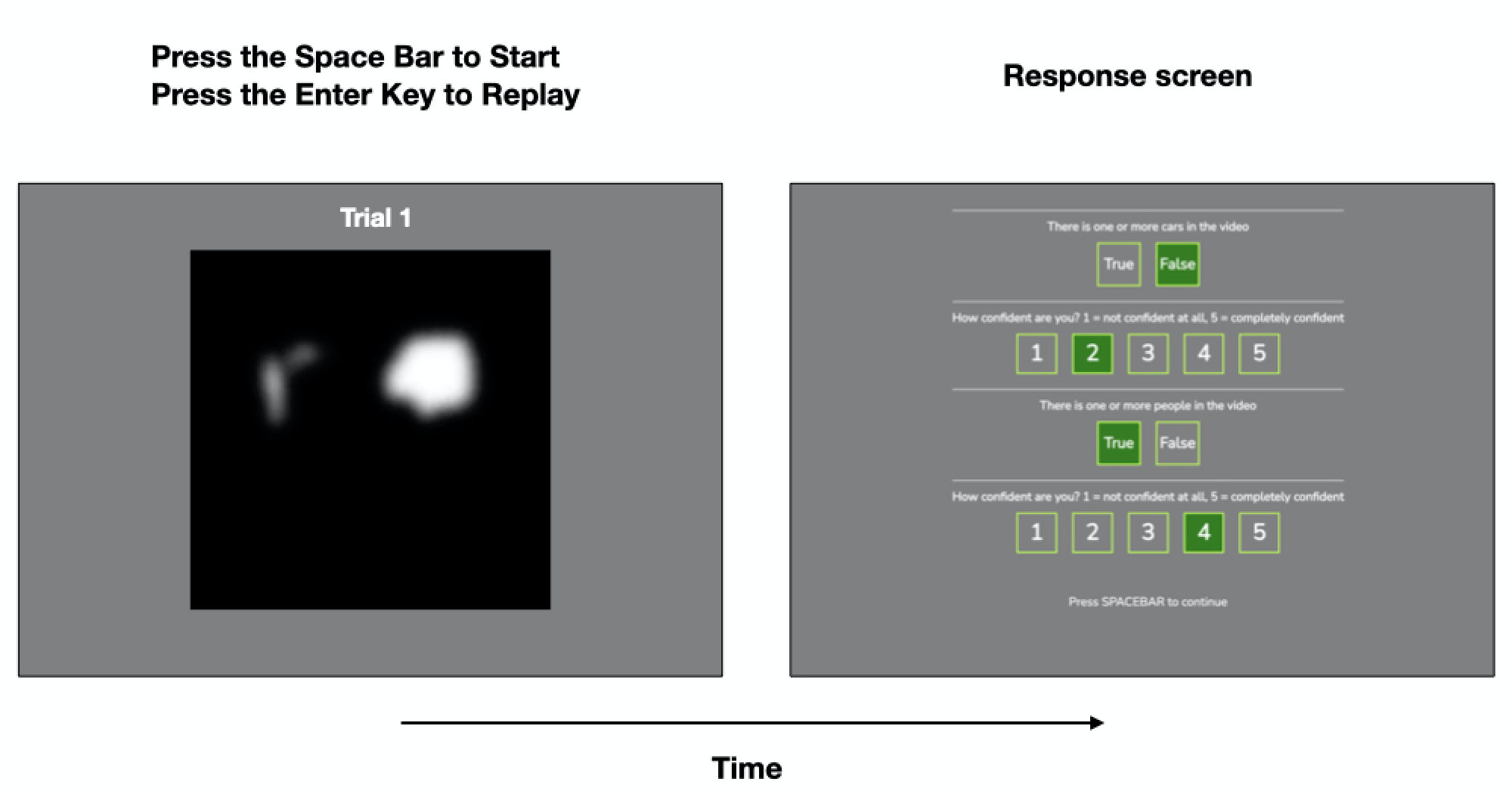}
    \caption{Example trial conducted using the SimplePhy online platform \cite{lago_miguel_simplephy_nodate}. After watching a five-second long video clip of a \acf{SPV} outdoor scene, participants had to indicate whether they believe cars and people to be present in the scene.
    Participants also indicated their confidence on a five-level Likert scale (1 = not confident at all, 5 = completely confident).}
    \label{fig:procedure_flow}
\end{figure*}

\subsection{Scene Simplification Strategies}
\label{sec:strategies}

Stimuli were processed by four different scene simplification strategies, adapted from state-of-the-art computer vision algorithms.

\subsubsection{Highlighting Visually Salient Information}
We used DeepGaze II \cite{kummerer_deepgaze_2016} to highlight visually salient information in the recorded videos. DeepGaze II produced a saliency map that assigned an importance value $\in [0,1]$ to each pixel in a given video frame.
Importance values were then linearly mapped to stimulus amplitudes applied to the simulated retinal implant.

\subsubsection{Substituting Depth for Intensity}
\label{sec:depth}
We used a self-supervised monocular depth estimation model called monodepth2 \cite{godard_digging_2019} (specifically the pre-trained \texttt{mono+stereo\_640x192 model}) to predict a per-pixel relative depth map from each frame in the videos.
We first sorted the depth values of all pixels in a frame and removed all depth values above the 80th percentile (where the 0th and 100th percentile referred to the nearest and farthest pixels to the viewer, respectively).
We then applied an exponential decay on the depth values such that the closest pixels had grayscale value 180 and the farthest pixels had grayscale value 0. 

\subsubsection{Object Segmentation}
\label{sec:segmentation}
To segment objects of interest from background clutter, we used a combination of the scene parsing algorithm from the MIT Scene Parsing Benchmark \cite{zhou_semantic_2016,zhou_scene_2017} and an object segmentation algorithm called detectron2 \cite{wu_yuxin_detectron2_2019}.
Given that all the stimuli were outdoor scenes, we obtained the detected object binary masks that were labeled as a person, bicycle, car, or bus for each video frame. If there was no object detected in the scene, then we only represented the main structural edges from the scene-parsing algorithm. The scene-parsing algorithm sometimes produces more than 50 parsed regions from the scene. In order to produce less clustered output, we only preserve the regions labeled as roads or sidewalks. For the parsed regions, we then extracted the structural edges for better visualization in the end.
The resulting binary masks were then linearly mapped to stimulus amplitudes applied to the simulated retinal implant.

\subsubsection{Combining Saliency, Depth, and Segmentation}
\label{sec:combination}
Recognizing the complementary qualities of the three algorithms described above, we wondered whether a combination of saliency, depth, and object segmentation could further improve scene understanding.
While segmentation excels at highlighting objects of interests, it might miss regions of interest that do not have a clear semantic label (which would be highlighted by the more bottom-up--driven saliency detector) or nearby obstacles (which would be highlighted by the depth algorithm).
To arrive at a binary mask of salient objects, we thresholded the saliency map to retain only the \SI{10}{\percent} most salient pixels and combined it with the object segmentation map using a logical OR.
We then scaled the grayscale value of each pixel in the new binary mask with a quadratic function of depth, similar to the above: $y = -\frac{45}{16}(\frac{8}{d_{\max}-d_{\min}}x-\frac{16}{d_{\max}-d_{\min}})^2 + 180$.

\subsection{Simulated Prosthetic Vision}
\label{sec:SPV}

The preprocessed videos were then used as input stimuli to the pulse2percept simulator \cite{beyeler_pulse2percept_2017}, which provides an open-source implementation of Ref.~\cite{beyeler_model_2019} (among others).
The simulator takes a downscaled version of the preprocessed image, and interprets the grayscale value of each pixel in a video frame as a current amplitude delivered to the simulated retinal implant.
However, pulse2percept describes the output of \ac{SPV} not as a pixelated image, but determines the shape of each phosphene based on the retinal location of the simulated implant as well as model parameters $\rho$ and $\lambda$ (see Section~\ref{sec:background}).
As can be seen in (Fig.~\ref{fig:axonmap}, \emph{left}), electrodes near the horizontal meridian were thus predicted to elicit circular percepts, while other electrodes were predicted to produce elongated percepts that differed in angle based on whether they fell above or below the horizontal meridian.

Importantly, $\rho$ and $\lambda$ seem to vary drastically across patients \cite{beyeler_model_2019}. Although the reason for this is not fully understood, it is clear that the choice of these parameter values may drastically affect the quality of the generated visual experience.
To cover a broad range of potential visual outcomes, we thus simulated nine different conditions with $\rho = \{100, 300, 500\}$ and $\lambda = \{0, 1000, 5000\}$.

To study the effect that the number of electrodes in a retinal implant has on scene understanding, we simulated three different retinal implants consisting of $8 \times 8$, $16 \times 16$, and $32 \times 32$ electrodes arranged on a rectangular grid.
These sizes roughly correspond to existing and near-future retinal implants.

\subsection{Virtual Patients}
\label{sec:virtual_patients}

We recruited 45 sighted undergraduate students (ages 18--21; 31 females, 14 males) from the student pool at \ac{UCSB} to act as virtual patients in our experiments.
Subjects were asked to watch \ac{SPV} videos depicting various outdoor scenes and indicate whether they believe people and/or cars to be present in the scene.
We were primarily interested in investigating their perceptual performance as a function of the four different scene simplification strategies, three retinal implant resolutions, and nine combinations of model parameters $\rho$ and $\lambda$.
All experiments were performed under a protocol approved by the university's Institutional Review Board.

\subsubsection{Experimental Setup and Apparatus}
The experiment was set up using a recent online platform called SimplePhy \cite{lago_miguel_simplephy_nodate}. All subjects completed the experiment online using a personal laptop or computer.

We used a between-subjects design where each subject was randomly assigned to one of the nine model parameter conditions ($\rho \in \{100, 300, 500\} \times \lambda \in \{0, 1000, 5000\}$).
Each condition was completed by five different subjects.
Within each condition, each subject completed all 16 videos with the four scene simplification strategies (depth, saliency, segmentation, combination) and three electrode grid resolutions ($8 \times 8$, $16 \times 16$, $32 \times 32$).
Therefore, each subject completed 192 trials ($16 \times 4 \times 3$) in total, which took about 45--60 minutes to finish.

\subsubsection{Experimental Task and Procedure}
Subjects underwent a short online practice session consisting of 8 practice trials, where they were shown original videos from the head-mounted camera alongside their corresponding \ac{SPV} videos.
An example trial is shown in Fig.~\ref{fig:procedure_flow}.
Note that the video sequences used in the practice session did not appear in the actual experiment.
After each video, a new screen appeared (`response screen' in Fig.~\ref{fig:procedure_flow}) on which subjects indicated whether they believed the scene contained any people or cars.
Subjects also indicated their confidence on a five-level Likert scale (1 = not confident at all, 5 = completely confident). Detecting cars and people is an essential component for orientation \& mobility. Increasing a patient's ability to detect and recognize moving objects may prevent them from dangerous situations in real-life scenarios.

\subsubsection{Evaluating performance}
Perceptual performance was assessed using the sensitivity index ($d'$, ``d-prime''), which is a dimensionless statistic from signal detection theory that can be used to measure a participant's perceptual sensitivity \cite{simpson_what_1973}:
\begin{equation}
    d' = Z(\textrm{hit rate}) - Z(\textrm{false discovery rate}),
\end{equation}
where the function $Z(p)$, with $p \in [0,1]$, is the inverse of the cumulative distribution function of the Gaussian distribution.
Here, the hit rate was calculated from the number of trials in which a participant correctly identified people or cars to be present, and the \ac{FDR} was calculated from the number of trials in which a participant indicated to see either people or cars, although none of them were present.
A higher $d'$ indicates better ability to discriminate between trials in which a target is present (signal) and trials in which a target is absent (noise). $d'=0$ indicates that a participant is performing at chance levels.

We used bootstrapping to test for statistical significance. Ten thousand bootstrap re-samples were used to estimate within-subject and between-subject differences. All $p$ values were corrected using \ac{FDR} to control the probability of incorrectly rejecting the null hypotheses when doing multiple comparisons \cite{li_controlling_2015}.

For the sake of completion, perceptual performance was also evaluated on four common statistical indicators: accuracy (number of correct predictions), precision (number of correct predictions divided by the number of all trials containing either people or cars), recall (number of correct predictions divided by the number of all trials that should have been identified as containing either people or cars), and the F1 score (harmonic mean of the precision and recall).
Note that some of these are part of $d'$.


\section{Results}


\subsection{Effect of Scene Simplification Strategy on Perceptual Performance}
\begin{figure}[!t]
\centering
\includegraphics[width=0.9\columnwidth]{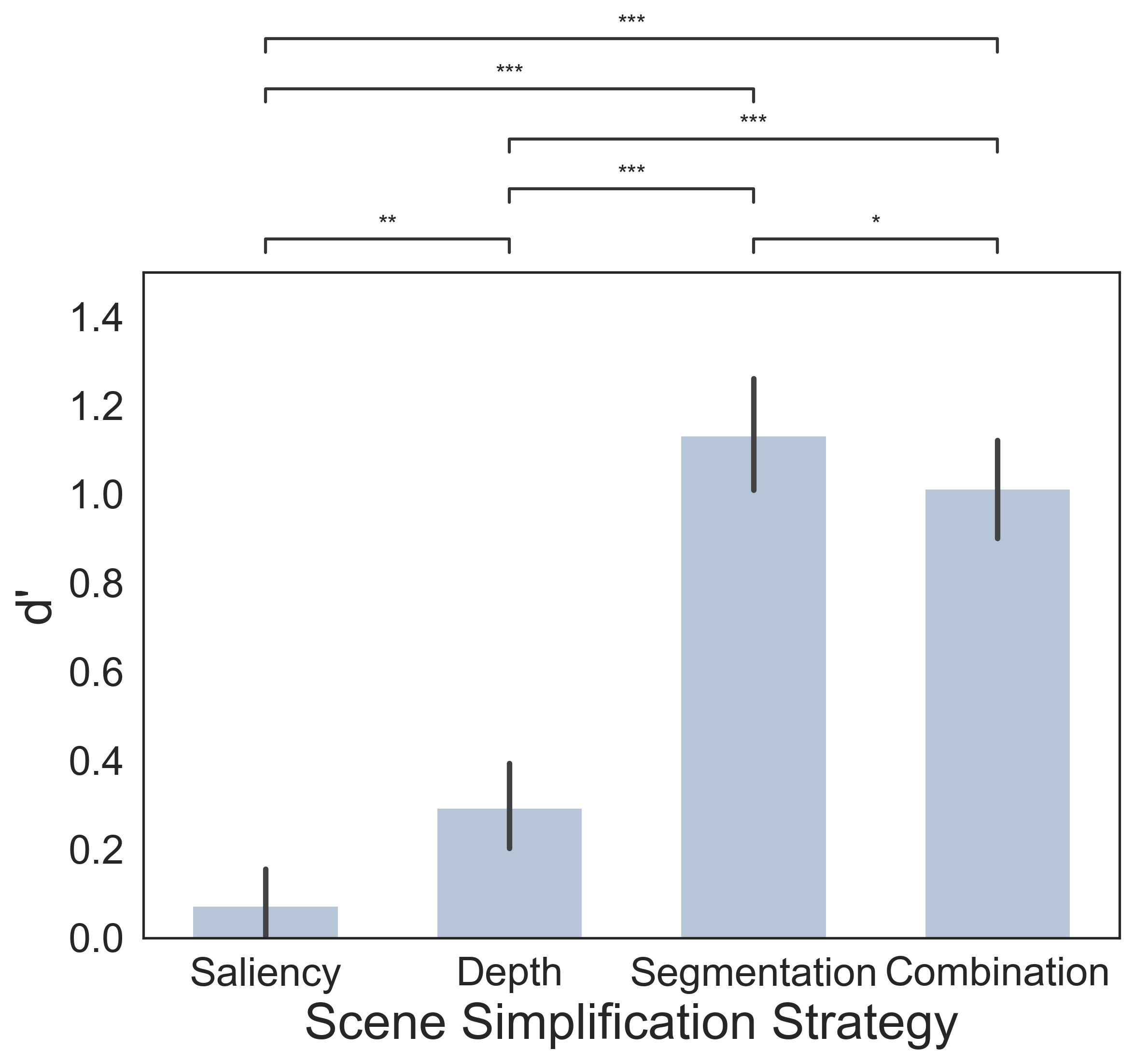}
\caption{Effect of Scene Simplification Strategy. The p values are corrected using \acf{FDR}, *<.05, **<.01, ***<.001. The error bars show the 95$\%$ confidence interval.}
\label{fig:results_algorithms}
\end{figure}

Fig.~\ref{fig:results_algorithms} shows the perceptual performance of virtual patients as a function of the different scene simplification strategies.
Subjects performed best using the segmentation algorithm ($d': \mu=1.13, \sigma=1.01$).
Performance based on saliency ($d'$: $\mu=0.07$, standard deviation $\sigma=0.66, p<0.001$), depth ($d': \mu=0.29, \sigma=0.66, p<0.001$), and combination  ($d': \mu=1.01, \sigma=0.91, p<0.05$) was significantly worse.
Saliency performed worse, followed by depth ($p<0.01$) and the combination algorithm ($p<0.001$).
Interestingly, the combination algorithm was not able to benefit from the complementary information contributed by the individual saliency, depth, and segmentation algorithms.

These findings are further corroborated by other objective measures such as accuracy, precision, and recall (see Table~\ref{tab:results_algorithms}) that reveal object segmentation as the most beneficial scene simplification strategy.

Subjects' confidence ratings in the segmentation condition ($\mu=3.02, \sigma=1.10$) and combination condition ($\mu=2.96, \sigma=1.08$) were both significantly higher than the those in the saliency condition ($\mu=2.65, \sigma=1.12$) and the depth condition ($\mu=2.68 \sigma=1.07$; all $p<0.001$). No difference between saliency and depth condition was found ($p=0.09$).

\begin{table}[!t]
\centering
\begin{tabular}{|r|rrrr|}
	\hline
	Condition & Accuracy & Precision & Recall & F1 \\
	\hline\hline
	Saliency & 0.51 & 0.53 & 0.46 & 0.46\\
	\hline
	Depth & 0.54 & 0.56 & 0.56 & 0.53\\
	\hline
	Segmentation & {\bf 0.68} & {\bf 0.73} & {\bf 0.63} & {\bf 0.68} \\
	\hline
	Combination & 0.66 & 0.72 &  0.62 & 0.67\\
	\hline
\end{tabular}
\caption{Virtual patient's ability to identify people and cars in outdoor scenes using different scene simplification strategies (bold: best overall).}
\label{tab:results_algorithms}
\end{table}

\subsection{Effect of Phosphene Size and Elongation on Perceptual Performance}
\begin{figure}[!tb]
\centering
\includegraphics[width=\columnwidth]{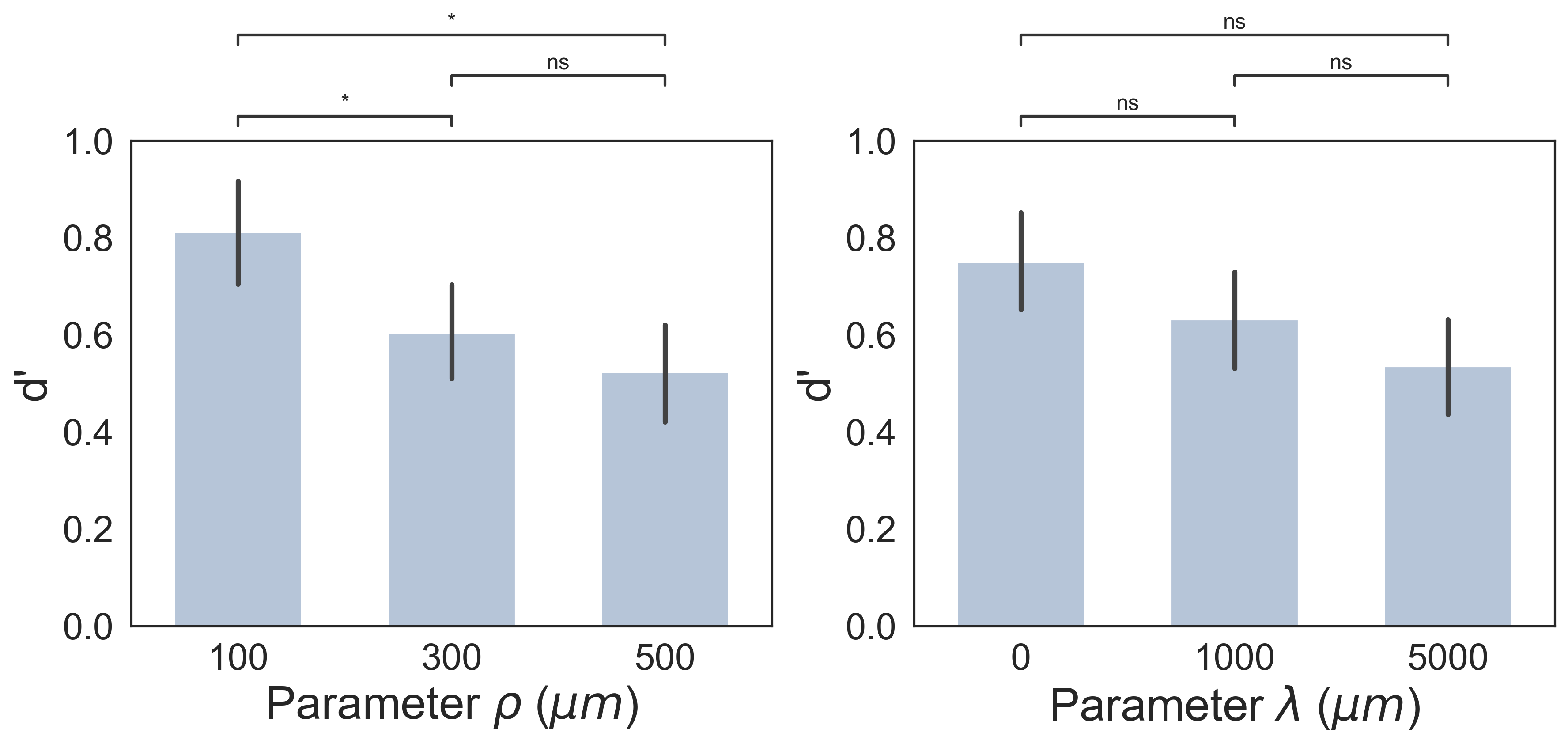}
\caption{Sensitivity index ($d'$) as a function of phosphene width ($\rho$) and length ($\lambda$). The $p$ values were corrected using \acf{FDR}: * $p<.05$, ** $p<.01$, *** $p<.001$. The error bars show the 95$\%$ confidence interval.}
\label{fig:results_rho_lambda}
\end{figure}

Fig.~\ref{fig:results_rho_lambda} shows the perceptual performance of virtual patients as a function of phosphene size ($\rho$) and elongation ($\lambda$).
As expected, smaller phosphenes led to better perceptual performance, with $\rho=\SI{100}{\micro\meter}$ scoring significantly better ($d': \mu=0.81, \sigma=1.02$) than $\rho=\SI{300}{\micro\meter}$ ($d': \mu=0.6, \sigma=0.89, p=0.03$) and $\rho=\SI{500}{\micro\meter}$ ($d': \mu=0.52, \sigma=0.96, p=0.02$).
No significant difference in $d'$ was found between the conditions with $\rho=\SI{300}{\micro\meter}$ and $\rho=\SI{500}{\micro\meter}$ ($p=0.28$).

\begin{figure}[!tb]
\centering
\includegraphics[width=\columnwidth]{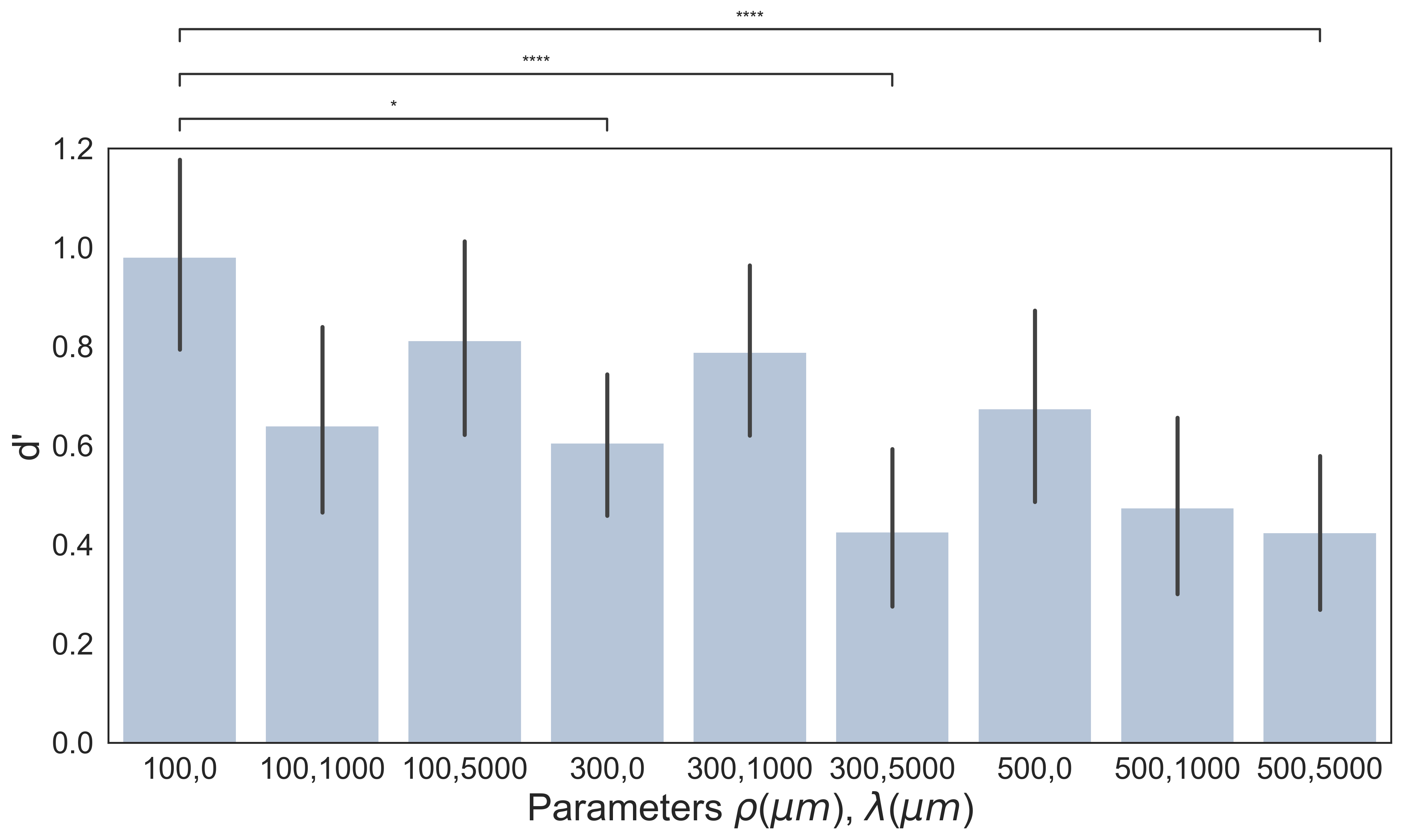}
\caption{Sensitivity index ($d'$) for each tested combination of phosphene width ($\rho$) and length ($\lambda$).The p values are corrected using \acf{FDR}, *<.05, **<.01, ***<.001. The error bars show the 95$\%$ confidence interval.}
\label{fig:results_all_rho_lambda}
\end{figure}

\begin{table}[!tb]
\centering
\begin{tabular}{|rr|rrrr|}
    \hline
	$\rho$ (\SI{}{\micro\meter}) & $\lambda$ (\SI{}{\micro\meter}) &  Accuracy & Precision & Recall & F1 \\
	\hline \hline
	100 & 0 & \textbf{\textit{0.64}} & \textbf{\textit{0.69}} & 0.53 & 0.56\\
	& 1000 & 0.59 & 0.58 & \textbf{\textit{0.70}} & \textbf{\textit{0.62}} \\
	& 5000 & 0.62 & 0.68 & 0.63 & 0.60\\
	\hline
	300 & 0 & 0.59 & 0.61 &  \textbf{0.62} & \textbf{0.58} \\
	& 1000 & \textbf{0.62} & \textbf{0.63} & 0.58 & \textbf{0.58} \\
	& 5000 & 0.57 & 0.58 & 0.56 & 0.55\\
	\hline
	500 & 0 & \textbf{0.60} & \textbf{0.60} &  0.63 & \textbf{0.59} \\
	& 1000 &  0.58 & 0.56 & \textbf{0.66} &  \textbf{0.59} \\
	& 5000 & 0.57 & 0.55 & \textbf{0.66} &  \textbf{0.59} \\
	\hline
\end{tabular}
\caption{Virtual patient's ability to identify people and cars in outdoor scenes as a function of phosphene size ($\rho$) and elongation ($\lambda$; bold: best in box, italics: best overall).}
\label{tab:results_all_rho_lambda}
\end{table}

A similar trend was evident with respect to phosphene elongation.
Here, $\lambda=0$ indicated circular phosphenes, similar to the \ac{SPV} studies described in Section~\ref{sec:related_work}, and led to similar performance as $\rho=\SI{100}{\micro\meter}$ ($d': \mu=0.75, \sigma=0.99$).
And while there was a trend evident indicating that more elongated phosphenes may lead to poorer perceptual performance, this trend did not reach statistical significance ($\lambda=\SI{1000}{\micro\meter}, d': \mu=0.63, \sigma=0.98, p>0.05$; $\lambda=\SI{5000}{\micro\meter}, d': \mu=0.53, \sigma=0.91, p>0.05$).

This trend could later be confirmed by investigating $d'$ across all nine model parameter conditions (see Fig.~\ref{fig:results_all_rho_lambda}).
Here we found a clear decreasing trend in $d'$ as phosphene size($\rho$) and phosphene elongation($\lambda$) increased.
However, notice that $d'$ was positive in all conditions, indicating that subjects performed better than chance even when phosphenes were unusually large and elongated.

Similar patterns were found in all the other behavioral performance measurements (see Table~\ref{tab:results_all_rho_lambda}).
Overall, the highest accuracy, precision, recall, and F1 scores (italics) were achieved with the smallest tested phosphene size ($\rho=\SI{100}{\micro\meter}$), but not necessarily with the shortest phosphene length. 

Unfortunately, the subjects' confidence ratings across conditions with different phosphene sizes ($\rho$) and phosphene elongations ($\lambda$) did not show any significant difference ($\rho=100: \mu=2.74, \sigma=1.19$; $\rho=300: \mu=2.95, \sigma=1.07$; $\rho=500: \mu=2.80, \sigma=1.04$; $\lambda=0: \mu=2.60, \sigma=1.14$; $\lambda=1000: \mu=3.18, \sigma=1.14$; $\lambda=5000: \mu=2.71, \sigma=.93$, all $p>.05$).

\begin{figure}[!t]
\centering
\includegraphics[width=.8\columnwidth]{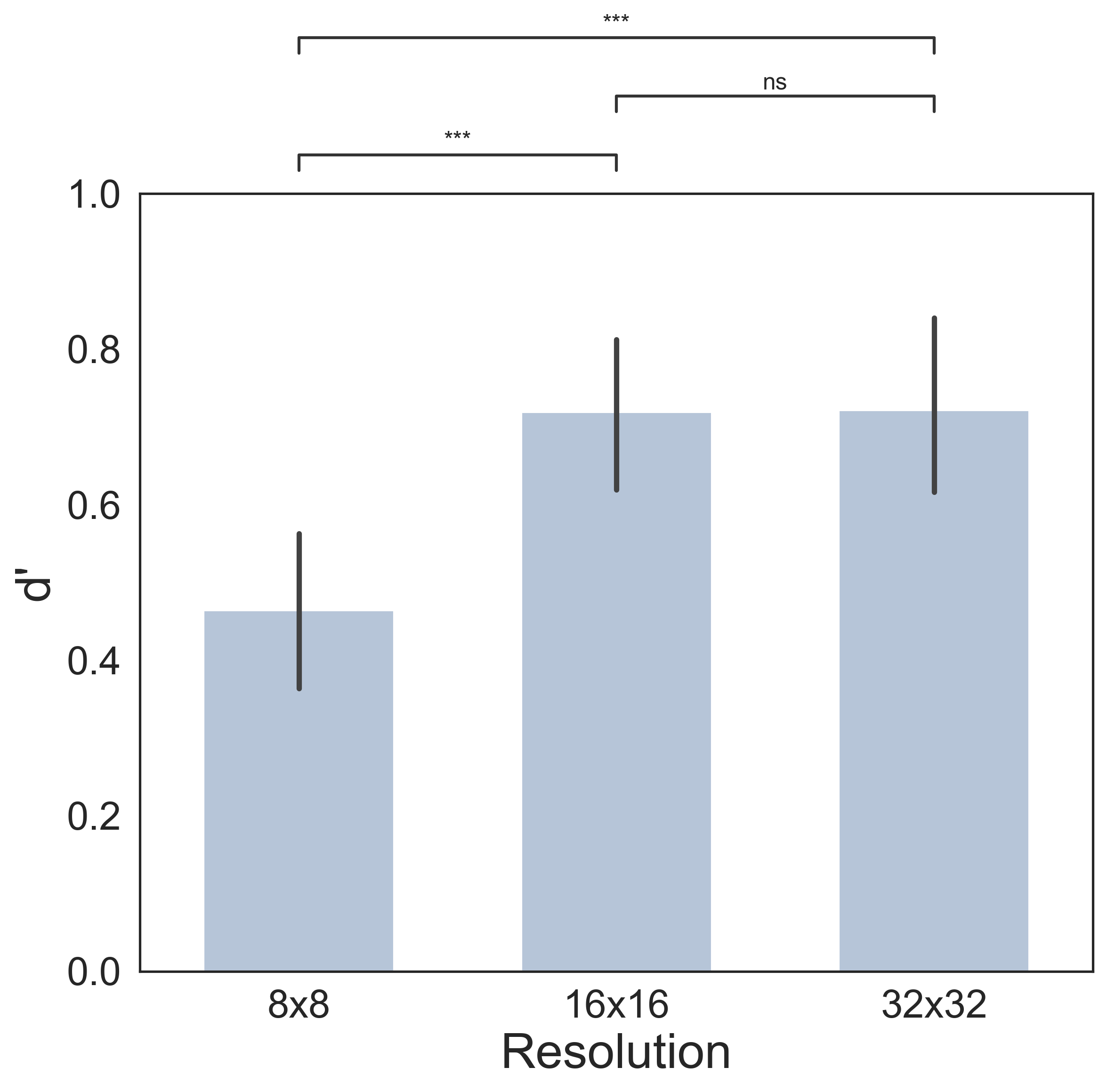}
\caption{Sensitivity index ($d'$) as a function of electrode grid size. The p values are corrected using \acf{FDR}, *<.05, **<.01, ***<.001. The error bars show the 95$\%$ confidence interval.}
\label{fig:results_resolution}
\end{figure}

\begin{table}[!t]
\centering
\begin{tabular}{|r|rrrr|}
	\hline
	Resolution & Accuracy & Precision & Recall & F1 \\
	\hline\hline
	8$\times$8 & 0.57 & 0.59 & 0.57 & 0.54\\
	\hline
	16$\times$16 & {\bf 0.61} & {\bf 0.62} & 0.64 & {\bf 0.61} \\
	\hline
	32$\times$32 & {\bf 0.61} & {\bf 0.62} & {\bf 0.65} & {\bf 0.61} \\
	\hline
\end{tabular}
\caption{Virtual patient's ability to identify people and cars in outdoor scenes as a function of electrode grid size (bold: best overall).}
\label{tab:results_resolution}
\end{table}

\subsection{Effect of Electrode Grid Size on Perceptual Performance} \label{effect_of_resolution}

Fig.~\ref{fig:results_resolution} shows the perceptual performance of virtual patients as a function of electrode grid size.
As expected, performance improved as the number of electrodes in the array was increased from $8 \times 8$ ($d': \mu=0.47, \sigma=0.87$) to $16 \times 16$ ($d': \mu=0.72, \sigma=0.93, p<0.001$).
However, further increasing the number of electrodes to $32 \times 32$ did not measurably affect performance ($p=0.37$).

This finding is again corroborated by accuracy, precision, recall, and F1 scores (Table~\ref{tab:results_resolution}), indicating virtually identical performance for $16 \times 16$ and $32 \times 32$.

Again, no significant difference in confidence ratings was found for different electrode array resolution ($8\times8: \mu=2.74, \sigma=1.12; 16\times16: \mu=2.85, \sigma=1.09; 32\times32:\mu=2.89, \sigma=1.10$, all $p>0.05$).

\section{Discussion}

\subsection{Object Segmentation May Support Scene Understanding}
The present study provides the first side-by-side comparison of several deep learning--based scene simplification strategies for bionic vision.
Considering a number of relevant implant configurations in combination with a psychophysically validated model of \ac{SPV}, we identified object segmentation as the most promising image processing strategy to support outdoor scene understanding of virtual patients (see Fig.~\ref{fig:results_algorithms} and Table~\ref{tab:results_algorithms}).
This finding is consistent with recent studies indicating the potential utility of semantic segmentation for bionic vision \cite{horne_semantic_2016,sanchez-garcia_indoor_2019}.

Object segmentation compared favorably with two other scene simplification strategies: based on visual saliency and monocular depth estimation.
Whereas the saliency model struggled with the lighting conditions of the outdoor data set (often highlighting regions of increased contrast, and falling victim to shadows), the depth model often failed to highlight nearby obstacles.
However, these models may prove their value in less structured test environments, where performance is less focused on semantic labeling and more concerned with the decluttering of complex scenes or the avoidance of nearby obstacles.

\subsection{Increased Phosphene Size Impedes Perceptual Performance}
To the best of our knowledge, this study is also the first to study \ac{SPV} with a neurobiologically inspired, psychophysically validated model of phosphene shape \cite{beyeler_model_2019}.
Whereas previous studies assumed that phosphenes are isolated, focal spots of light \cite{dagnelie_real_2007,chen_simulating_2009,lui_transformative_2011,perez-yus_depth_2017,sanchez-garcia_indoor_2019}, here we systematically evaluated perceptual performance across a wide range of common phosphene sizes ($\rho$) and elongation ($\lambda$).
As expected, the best performance was achieved with small, circular phosphenes ($\rho=\SI{100}{\micro\meter}, \lambda=0$; see Fig.~\ref{fig:results_rho_lambda}), and increasing phosphene size and elongation negatively affected performance (Fig.~\ref{fig:results_all_rho_lambda}).
This finding suggests that future studies of \ac{SPV} should take into account realistic phosphene shape when making predictions and drawing conclusions.

However, it is worth mentioning that the sensitivity index ($d'$) remained positive in all tested conditions, indicating that subjects performed better than chance even when phosphenes were unusually large and elongated.
This result suggests that all tested scene simplification strategies enabled the virtual patients to perform above chance levels, no matter how degraded the \ac{SPV} quality.


\subsection{Increasing the Number of Electrodes Does Not Necessarily Improve Performance}
As expected, perceptual performance improved as the size of the electrode grid was increased from $8 \times 8$ to $16 \times 16$. However, further increasing the number of electrodes to $32 \times 32$ did not measurably affect performance.
This result is consistent with previous literature suggesting that number of electrodes is not the limiting factor in retinal implants \cite{behrend_resolution_2011,beyeler_learning_2017}.

\subsection{Limitations and Future Work}

Although the present results demonstrate the utility of deep learning--based scene simplification for bionic vision, there are a number of limitations that should be addressed in future work.

First, in an effort to focus on scenes important for orientation and mobility, we limited our dataset to outdoor scenes.
However, it would also be valuable to evaluate the performance of different scene simplification strategies on indoor scenarios. Because indoor scenes have different layouts and types of objects, the algorithms studied here might have different performances compared to outdoor scenes. For example, the saliency model might perform better in highlighting salient regions without the interference of light and shadow contrasts.

Second, to keep the perceptual judgments amenable to quantitative performance measures, we limited the current study to a simple detection task involving common semantic object categories (i.e., people and cars). 
This might explain the superior performance of the semantic segmentation algorithm, which operates with semantic labels. In contrast, the depth and saliency algorithms might prove more valuable when applied to open-ended navigation tasks.
In the future, we plan to conduct such \ac{SPV} studies in immersive \ac{VR} to gain more comprehensive insight into the behavioral performance of virtual patients. 

Third, the present study should be understood as a first step towards the ultimate goal of creating a retinal implant supported by deep learning--based image preprocessing. Such a device would require all processing to happen in real time at the edge.
One solution could come in the form of low-power, low-latency neuromorphic hardware coupled with an event-based vision sensor.
Future iterations of this work may include end-to-end training of scene simplification strategies fitted to a specific implant technology or even an individual patient. Overall this work has the potential to drastically improve the utility of prosthetic vision for people blinded from retinal degenerative diseases.

\section{Data Availability}
\label{sec:data_availability}

All raw video sequences (original and preprocessed) are available on the Open Science Framework (\url{https://osf.io/s2udz}). \Ac{SPV} models were based on the pulse2percept Python package \cite{beyeler_pulse2percept_2017}. Code used to implement the scene simplification strategies is available on GitHub (\url{https://github.com/bionicvisionlab/2021-han-scene-simplification}, v0.1).

\section*{Acknowledgments}
This work was partially supported by the National Institutes of Health (NIH R00 EY-029329 to MB). We would like to thank Yaoyi Bai and Sikun Lin for their contribution to an earlier version of the depth algorithm, and Asa Young for collecting the video stimuli. We would also like to thank Dr. Miguel Lago for technical support with regards to the SimplePhy \cite{lago_miguel_simplephy_nodate} online platform.

\bibliographystyle{ACM_Reference_Format}
\bibliography{references}








\end{document}